\def\year{2021}\relax
\documentclass[letterpaper]{article} % DO NOT CHANGE THIS
\usepackage{aaai21}  % DO NOT CHANGE THIS
\usepackage{times}  % DO NOT CHANGE THIS
\usepackage{helvet} % DO NOT CHANGE THIS
\usepackage{courier}  % DO NOT CHANGE THIS
\usepackage[hyphens]{url}  % DO NOT CHANGE THIS
\usepackage{graphicx} % DO NOT CHANGE THIS
\usepackage{natbib}  % DO NOT CHANGE THIS AND DO NOT ADD ANY OPTIONS TO IT
\usepackage{caption} % DO NOT CHANGE THIS AND DO NOT ADD ANY OPTIONS TO IT
\usepackage{fancyhdr}
\usepackage{multirow}
\usepackage{subcaption}
\usepackage{sidecap}

\title{On the Importance of Word Order Information in Cross-lingual \\ Sequence Labeling}
\author{Zihan Liu, Genta Indra Winata, Samuel Cahyawijaya, \\ Andrea Madotto, Zhaojiang Lin, Pascale Fung \\}

\affiliations{
Center for Artificial Intelligence Research (CAiRE)\\
  The Hong Kong University of Science and Technology, Clear Water Bay, Hong Kong\\
  \texttt{zihan.liu@connect.ust.hk, pascale@ece.ust.hk}
 }

\begin{document}

\maketitle

\begin{abstract}
% Intuition:
% lower the effect of word information from sequence labeling.
% make model get less word order information
% Most of the recent works in cross-lingual adaptation do not consider the word order variances in different languages.
Cross-lingual models trained on source language tasks possess the capability to transfer to target languages directly. However, since word order variances generally exist in different languages, cross-lingual models that overfit into the word order of the source language could have sub-optimal performance in target languages.
% To verify this hypothesis, we investigate whether 
% fitting models with less source language word order information 
In this paper, we hypothesize that reducing the word order information fitted into the models can improve the adaptation performance in target languages.
% making models insensitive to the word order of the source language can improve the adaptation performance in target languages.
% we compare state-of-the-art sequence encoders, such as LSTM and Transformer, with our proposed methods, which fit less word order information for cross-lingual sequence labeling tasks.
% To do so, we introduce three methods to make models insensitive to word orders and compare them to order-sensitive models.
% propose to construct sequence encoding models that are less sensitive to word orders for cross-lingual sequence labeling tasks.
To verify this hypothesis, we introduce several methods to make models encode less word order information of the source language and test them based on cross-lingual word embeddings and the pre-trained multilingual model.
% and observe the performance changes.
% and compare them to order-sensitive models.
% In addition, based on this hypothesis, we propose a new method for fine-tuning multilingual BERT in downstream cross-lingual sequence labeling tasks.
Experimental results on three sequence labeling tasks (i.e., part-of-speech tagging, named entity recognition, and slot filling tasks) show that reducing word order information injected into the model can achieve better zero-shot cross-lingual performance. Further analysis illustrates that fitting excessive or insufficient word order information into the model results in inferior cross-lingual performance. Moreover, our proposed methods can also be applied to strong cross-lingual models and further improve their performance.
% \footnote{The code is attached in the supplementary material.}
% performance vs. language distance
% Additional experiments: such as Few-shot
\end{abstract}

\section{Introduction}
% 1. Introduce Cross-lingual Adaptation. Word order variances is less studied
Neural-based data-driven supervised approaches have achieved remarkable performance in sequence labeling tasks (e.g., named entity recognition)~\cite{lample2016neural, devlin2019bert}. However, these methods are not applicable to low-resource languages, where extensive training data are absent. Recently, numerous cross-lingual adaptation methods have been applied to this data-scarcity scenario, where zero or very few target language training samples are utilized~\cite{wisniewski2014cross,schuster2019cross,artetxe2019massively,liu2019zero,chen2019multi}.

% However, most of the cross-lingual research works ignore the word order differences across languages and utilize 
% However, there has been less research done in terms of the word order differences across languages. 
Despite the focus on cross-lingual methods, the word order differences across languages is a less studied problem of the cross-lingual task.
For cross-lingual models, sequence encoders that are based on LSTM~\cite{hochreiter1997long} or Transformer~\cite{vaswani2017attention} inevitably model the word order information in the source language~\cite{xie2018neural,liu2019zero}, which we characterize the \textit{order-sensitive} property~\cite{ahmad2018difficulties}. Since different languages have different word orders, models that fit into the word order of the source language could hurt the performance in the target languages. 
% Therefore, in this paper, we investigate whether fitting models with less source language word order information can improve the adaptation performance in target languages.

% Therefore, in this paper, we propose to build order-reduced models which encode less word order information compared to existing sequence encoders, such as LSTM and Transformer.
In this paper, we hypothesize that making models less sensitive to word orders can boost cross-lingual performance, and then we introduce four methods to construct \textit{order-reduced} models to verify this hypothesis.
% To do so, we propose three methods to make construct \textit{order-reduced} models.
% , and we compare them with the baseline models.
First, we propose an Order-Reduced Transformer (ORT), which removes the positional embeddings from Transformer and utilizes one-dimensional convolutional networks to replace the linear layer as the feed-forward layer to encode partial order information; thus, the model becomes less dependent on the word order.
Second, we permutate the word order of the training samples, and order-sensitive models trained with them become insensitive to the word order.
% make models robust to order differences across languages.
Third, we hypothesize that the positional embeddings in multilingual BERT (M-BERT)~\cite{devlin2019bert} are order-agnostic given the surprising cross-lingual ability that it has~\cite{wu2019beto}.
Hence, we take the positional embeddings from M-BERT to initialize the positional embeddings in Transformer, and we freeze them in the training phase to make the model order-agnostic.
Additionally, based on the third method, we propose to freeze the positional embeddings when we fine-tune M-BERT to downstream cross-lingual tasks, which makes the model avoid fitting into the word order of the source language.

% Moreover, for sequence labeling tasks, conditional random field (CRF), which models the conditional probability of label sequences, could also implicitly model the source language word order in training. Therefore, we study whether removing the CRF layer helps improve cross-lingual performance.

% We conduct experiments based on cross-lingual word embeddings RCSLS~\cite{joulin2018loss} and multilingual BERT~\cite{devlin2019bert}.

% Experimental results show that order-reduced models are more robust to shuffled sequences and outperforms the baseline models for zero-shot cross-lingual sequence labeling tasks, namely, part-of-speech tagging (POS), named entity recognition (NER), and dialogue natural language understanding (NLU). In addition, our methods can be applied to competitive cross-lingual models and improve their performance.
We conduct experiments on zero-shot cross-lingual sequence labeling tasks, namely, part-of-speech tagging (POS), named entity recognition (NER), and slot filling (SF),\footnote{SF is a critical task in natural language understanding (NLU) for dialog systems.}, and we compare our models with order-sensitive sequence encoders, such as LSTM and Transformer.
% We summarize our contributions as follows:
% \begin{itemize}
%     \item Our proposed methods are robust to the shuffled sequences and bring consistent better zero-shot cross-lingual performance to competitive baseline models.
%     \item We introduce a new method for fine-tuning M-BERT to cross-lingual sequence labeling tasks, which achieves better performance than the original fine-tune method.
%     \item We conduct experiments in the few-shot setting to analyze the relations between the performance improvement of our models and the number of target language training samples.
% \end{itemize}
% From the experimental results, 
We summarize our insights as follows:
\begin{itemize}
    \item Order-reduced models are robust to word order shuffled sequences and consistently outperform order-sensitive models, including the state-of-the-art model.
    \item Retaining the order-agnostic property of M-BERT positional embeddings gives a better capability to generalize to target languages.
    % \item Overfitting the model to the word order information of the source language or removing the model dependency to the word order information (i.e., the most insensitive to word order) would badly impact to the target language performance.
    \item Encoding excessive or insufficient word order information leads to sub-optimal cross-lingual performance, and models that do not encode any word order information (i.e., the most insensitive to word order) give a poor performance on both source and target languages.
    % Encoding at least partial word order information is necessary. 
    % Models that do not encode any word order information (i.e., the most insensitive to word order) perform poorly on both source and target languages.
\end{itemize}
% 1. order-reduced are robust to the shuffled sequences and consistently outperforms order-sensitive models on three downstream tasks in all languages
% 2. Retaining pre-trained positional embedding in M-BERT is useful (two experiments: fine-tune M-BERT and replaced PE with M-BERT PE)
% 3. If we remove all word order, ... We found out that the information from neighbor words is useful. And CNN works better than Linear.

\section{Related Work}
% 1. Low-resource scenario. Cross-lingual adaptation
\subsection{Cross-lingual Adaptation}
% Coping with the scenario where zero or very few training samples are available is always an interesting and challenging research topic~\cite{gu2018universal}. 
Recently, cross-lingual sequence labeling methods that circumvent the need for extensive training data in target languages have achieved remarkable performance~\cite{kim2015part,kim2017cross,ni2017weakly,mayhew2017cheap,liu2020attention,liu2020cross,huck2019cross}. \citet{chen2019multi} 
% proposed mixture-of-experts models to 
utilized the similarity between the target language and each individual source language to achieve promising results on the cross-lingual NER task, while \citet{liu2019zero} utilized task-related keywords to build robust cross-lingual natural language understanding (NLU) systems. Taking this further, cross-lingual language models~\cite{pires2019multilingual,lample2019cross,huang2019unicoder,conneau2019unsupervised,liang2020xglue} pre-trained on a large data corpus achieved the state-of-the-art performance in multiple cross-lingual adaptation tasks.
% , such as NER and POS.

\subsection{Coping with Word Order Differences}
% 2. Word order
Word order differences across languages have been considered in cross-lingual dependency parsing~\cite{tiedemann2016synthetic, zhang2019cross} by using Treebank translation. For the same task, \citet{ahmad2018difficulties}, on the other hand, leveraged a relative positional self-attention encoder~\cite{shaw2018self} to make the sequence encoder less sensitive to word order and increase the adaptation robustness for target languages that are topologically different from the source language. Compared to the previous works, we conduct extensive experiments and analyses to illustrate the effectiveness of order-reduced models for cross-lingual sequence labeling tasks.
% , and our methods do not require an external library, such as Treebank.

\section{Methodology}
% First part:
%   1. Make positional embeddings less effectiveness
%   2. Order shuffled training samples
%   3. Multilingual BERT positional embeddings
% Second part:
%   Fine-tune M-BERT methods
% Third part:
%   Discussion about conditional random field
In this section, we introduce the proposed methods to reduce the word order of the source language fitted to order-sensitive sequence encoders.
% labeling models.
%, which can be split into two subsections.
% First, we introduce the methods to make models insensitive to word orders. Second, we introduce a new method for fine-tuning M-BERT to downstream cross-lingual sequence labeling tasks.
% Third, we discuss the effects of the conditional random field (CRF) for the sequence labeling tasks.

% \subsection{order-reduced Models}  \label{oi-models}
% % Our sequence encoding models are based on the Transformer encoder~\cite{vaswani2017attention}, and we propose three methods to construct order-reduced models.
% Three methods are introduced in this subsection to make model 

\subsection{Order-Reduced Transformer}
Given that Transformer~\cite{vaswani2017attention} relies on positional embeddings to encode word order information, we propose to remove them so as to reduce the word order information injected into input sentences. Note that given a linear layer as the feed-forward layer for Transformer, as introduced in~\citet{vaswani2017attention}, removing the positional embeddings module would mean getting rid of all the word order information, which would lead to a large performance drop in the source language and low performance for the cross-lingual transfer. Therefore, we utilize the one-dimensional convolutional network (Conv1d)~\cite{kim2014convolutional} as the feed-forward layer to extract n-gram features from the Multi-Head Attention features. Specifically, we formulate the encoding process as follows:
\begin{equation}
    g[1:n] = \texttt{MultiHead}(E(X[1:n])),
\end{equation}
where $X[1:n]$ represents the $n$-token input sequence; $E$ denotes the embedding layer; and $g[1:n] \in R^{n \times d} $, where $d$ is the hidden size of Transformer, represents the sequence features generated by Multi-Head Attention.
% The convolution operation involves a filter $\textbf{w} \in R^{hd}$, where h is the kernel size, and d is the feature dimension from the MultiHead Attention, and \textbf{w} is applied to a window of h features to produce a new feature. 
% For example, a feature $c_{i}$ is generated from a window of features $\text{g}_{i:i+h-1}$ by
% \begin{equation}
%     c_{i} = f(\textbf{w} \cdot \text{g}_{i:i+h-1} + b),
% \end{equation}
% where $b \in R$ is a bias term and f is a non-linear function. We add padding for this convolution process to ensure the output feature length is the same as the length of input tokens.

After that, a feature $c_i$ is generated from the window of features $g[i:i+h-1]$ by
\begin{equation}
  c_i = \texttt{Conv1d}(g[i:i+h-1]),
\end{equation}
where $h$ is the kernel size of Conv1d and the dimension of $c_i$ is equal to the number of output channels in Conv1d. 
We add padding for this convolution process to ensure the output feature length is the same as the length of the input tokens. Finally, the output feature sequences from Conv1d are the concatenation of $c_i$, where $i \in [1,n]$.

In this way, we fit the model with less word order information since the model only encodes the local n-gram features, and the prediction for each token is made based on the token itself and its neighbor tokens.

\subsection{Shuffling Word Order} \label{shuffle_sec}
% To fit models with less word order information and make them robust to the order differences across languages, 
Instead of removing positional embeddings, we propose to permutate the word order of input sequences in the source language training samples so as to train models to be robust for different word orders.
% To do so, we add perturbations to the word order of input sequences in the source language training samples, 
Meanwhile, we keep the order of tokens in each entity the same and consider them as one ``word'' to ensure we don't break entities in the sequences.

We follow~\citet{lample2017unsupervised} to generate permutations similar to the noise observed with word-by-word translation (i.e., word order differences across languages). Concretely, we apply a random permutation $\sigma$ to the input sequence, verifying the condition $\forall i \in \{1,n\}, |\sigma(i) - i| \leq k$, where n is the length of the input sentence and k is a tunable parameter that controls the shuffling degree.
% In our experiments, we set $k=\infty$.
% we generate a random vector q of size n, where $q_i=i+U(0,\alpha)$, and U is a draw from the uniform distribution in the specified range.
% To make sure the tokens
% Concretely, we randomly shuffle the word order for each sequence, while we keep the order of tokens in each entity the same. 
We use the order-shuffled training samples to train the models and make them less sensitive to word orders. 

\subsection{Order-Agnostic Positional Embeddings}
Another alternative method is to make the positional embeddings of Transformer order-agnostic to encode less order information.
% so that the models can still encode less order information. 
In light of M-BERT's astonishing cross-lingual performance~\cite{pires2019multilingual, wu2019beto},
we speculate that the positional embeddings in M-BERT are order-agnostic.
Hence, we leverage M-BERT's positional embeddings to initialize the positional embeddings for Transformer, and we freeze them in the training phase to prevent them from fitting into the source language word order.
In the experiments, since M-BERT's positional embeddings are based on its tokenizer, we leverage M-BERT's tokenizer to tokenize sequences and generate cross-lingual embeddings from M-BERT. Then, a Transformer encoder with M-BERT's positional embeddings is added on top of the M-BERT embeddings. We freeze the parameters of M-BERT in the training phase to ensure the cross-lingual embeddings from M-BERT do not fit into the source language word order.

\begin{SCtable*}
\renewcommand{\arraystretch}{1.1}
\centering
\resizebox{0.7\textwidth}{!}{
\begin{tabular}{lccccccccc}
\hline
\multicolumn{1}{c|}{\multirow{2}{*}{}} & \multicolumn{5}{c|}{\textbf{Named Entity Recognition Task}}                              & \multicolumn{4}{c}{\textbf{Slot Filling Task}}              \\ \cline{2-10} 
\multicolumn{1}{l|}{}                  & \textbf{en}    & \textbf{es}    & \textbf{nl}    & \multicolumn{1}{c|}{\textbf{de}}    & \multicolumn{1}{c|}{\textbf{avg}}   & \textbf{en}    & \textbf{es}    & \multicolumn{1}{c|}{\textbf{th}}    & \textbf{avg}   \\ \hline
\multicolumn{1}{l|}{Dist. to English}        & 0.00  & 0.12  & 0.14  & \multicolumn{1}{c|}{0.14}  & \multicolumn{1}{c|}{0.13}  & 0.00  & 0.12  & \multicolumn{1}{c|}{0.31}  & 0.22  \\ \hline \hline
\multicolumn{10}{l}{\textbf{\textit{Frozen Word-level Embeddings}}} \\ \hline
\multicolumn{1}{l|}{BiLSTM}            & 87.99 & 33.71 & 25.28 & \multicolumn{1}{c|}{15.28} & \multicolumn{1}{c|}{24.76} & \textbf{94.87} & 59.51 & \multicolumn{1}{c|}{20.63} & 40.07 \\
\multicolumn{1}{l|}{\hspace{3mm} w/ shuffled data}  & 83.85 & 30.09 & 22.87 & \multicolumn{1}{c|}{13.22} & \multicolumn{1}{c|}{22.06} & 93.57 & 62.02 & \multicolumn{1}{c|}{21.43} & 41.73 \\
\multicolumn{1}{l|}{Transformer (TRS)}       & \textbf{88.67} & 30.76 & 30.54 & \multicolumn{1}{c|}{18.53} & \multicolumn{1}{c|}{26.61} & 94.78 & 62.67 & \multicolumn{1}{c|}{22.33} & 42.50 \\
\multicolumn{1}{l|}{\hspace{3mm} w/ shuffled data}  & 82.75 & 28.54 & 28.43 & \multicolumn{1}{c|}{16.17} & \multicolumn{1}{c|}{24.38} & 92.07 & 63.86 & \multicolumn{1}{c|}{24.17} & 44.02 \\
\multicolumn{1}{l|}{\citet{ahmad2018difficulties}}       & 87.86 & 32.49 & 31.83 & \multicolumn{1}{c|}{19.24} & \multicolumn{1}{c|}{27.85} & 94.23 & 62.07 & \multicolumn{1}{c|}{23.14} & 42.61 \\ \hline
\multicolumn{1}{l|}{ORT}               & 88.41 & \textbf{34.33} & \textbf{33.54} & \multicolumn{1}{c|}{\textbf{24.14}} & \multicolumn{1}{c|}{\textbf{30.67}} & 94.50 & \textbf{66.84} & \multicolumn{1}{c|}{\textbf{25.53}} & \textbf{46.19} \\ \hline \hline
\multicolumn{10}{l}{\textbf{\textit{Frozen M-BERT Embeddings}}} \\ \hline
\multicolumn{1}{l|}{Transformer (TRS)}       & 89.53 & 58.93 & 46.28 & \multicolumn{1}{c|}{63.15} & \multicolumn{1}{c|}{56.12} & \textbf{94.93} & 46.75 & \multicolumn{1}{c|}{9.76}  & 28.26 \\
\multicolumn{1}{l|}{\hspace{3mm} w/ M-BERT PE}       & 88.44 & 58.27 & \textbf{47.63} & \multicolumn{1}{c|}{64.12} & \multicolumn{1}{c|}{56.67} & 94.53 & 47.23 & \multicolumn{1}{c|}{\textbf{10.06}} & 28.65 \\
\multicolumn{1}{l|}{\citet{ahmad2018difficulties}}       & \textbf{89.96} & \textbf{60.55} & 45.43 & \multicolumn{1}{c|}{61.58} & \multicolumn{1}{c|}{55.85} & 94.38 & 47.80 & \multicolumn{1}{c|}{8.83}  & 28.32 \\ \hline
\multicolumn{1}{l|}{ORT}               & 89.46 & 58.35 & 45.95 & \multicolumn{1}{c|}{\textbf{66.31}} & \multicolumn{1}{c|}{\textbf{56.87}} & 94.55 & \textbf{48.42} & \multicolumn{1}{c|}{9.92}  & \textbf{29.17} \\ \hline \hline
\multicolumn{10}{l}{\textbf{\textit{M-BERT Fine-tuning}}} \\ \hline
\multicolumn{1}{l|}{Fine-tune M-BERT}         & \textbf{91.95} & 74.49 & 69.13 & \multicolumn{1}{c|}{77.32} & \multicolumn{1}{c|}{73.65} & \textbf{95.97} & 69.41 & \multicolumn{1}{c|}{10.45} & 39.93 \\
\multicolumn{1}{l|}{\hspace{3mm} w/ frozen PE}      & 91.87 & \textbf{74.98} & \textbf{70.22} & \multicolumn{1}{c|}{\textbf{77.63}} & \multicolumn{1}{c|}{\textbf{74.28}} & 95.90 & \textbf{70.30} & \multicolumn{1}{c|}{\textbf{12.53}} & \textbf{41.42} \\ \hline
\end{tabular}
}
\caption{Zero-shot cross-lingual results on NER and SF tasks (averaged over three runs) for the three settings. 
% Except the M-BERT fine-tuning, the embeddings are frozen in the training stage to ensure the cross-lingual alignment is preserved.
We freeze the word-level embeddings in the training stage to ensure their cross-lingual alignment is preserved.
We use ``w/ shuffled data'' to denote the models trained with the word order shuffled source language training samples. ``PE'' denotes positional embeddings, and we use ``w/ M-BERT PE'' to represent that the model initialized with the frozen M-BERT positional embeddings. ``avg'' denotes the average performance over the target languages (English is excluded).}
\label{zero_shot_results}
\end{SCtable*}

\subsection{Fine-tuning M-BERT}
The original fine-tuning of M-BERT to downstream cross-lingual tasks is done by adding a linear layer on top of M-BERT and fine-tuning all the parameters of the model to the source language task~\cite{pires2019multilingual, wu2019beto}. This inescapably fits the model with the source language word order.
To circumvent this issue, we freeze the positional embeddings in M-BERT in the fine-tuning stage. By doing so, the positional embeddings can still provide the word order information for M-BERT to encode input sequences, and the model avoids fitting the word order of the source language.

% \subsection{Removing the CRF Layer}
% Combining sequence encoder such as bi-directional LSTM with conditional random field (CRF) has become a commonly used architecture for monolingual~\cite{lample2016neural,winata2019hierarchical} as well as cross-lingual sequence labeling tasks~\cite{xie2018neural,schuster2019crosslingual,liu2019attention}. 
% Since different languages have different word order patterns, the pattern for label sequences might be different as well.
% However, the CRF layer models the conditional probability of label sequences, which implicitly contains the word order information. Therefore, we try to remove the CRF layer, and the word-level predictions are obtained by a linear layer with softmax.

% \usepackage{multirow}
\begin{table*}[ht!]
\renewcommand{\arraystretch}{1.09}
\centering
\resizebox{0.99\textwidth}{!}{
\begin{tabular}{c|c|cccccc|cccc|cc}
\hline
\multirow{2}{*}{\textbf{Lang}} & \multirow{2}{*}{\begin{tabular}[c]{@{}c@{}}\textbf{Dist. to} \\ \textbf{English}\end{tabular}} & \multicolumn{6}{c|}{\textbf{Frozen Word-level Embeddings}}                                                                                                                           & \multicolumn{4}{c|}{\textbf{Frozen M-BERT Embeddings}}                                     & \multicolumn{2}{c}{\textbf{M-BERT Fine-tuning}}                                    \\ \cline{3-14} 
                      &                                                                              & BiLSTM & \begin{tabular}[c]{@{}c@{}}BiLSTM w/\\ shuffled data\end{tabular} & TRS   & \begin{tabular}[c]{@{}c@{}}TRS w/ \\ shuffled data\end{tabular} & RPT   & ORT   & TRS   & \begin{tabular}[c]{@{}c@{}}TRS w/ \\ M-BERT PE\end{tabular} & RPT   & ORT   & M-BERT & \begin{tabular}[c]{@{}c@{}}M-BERT w/ \\ frozen PE\end{tabular} \\ \hline
en                    & 0.00                                                                         & \textbf{93.76}  & 87.66                                                             & 93.07 & 86.84                                                           & 92.77 & 93.74 & 92.73 & 92.36                                                       & 92.47 & \textbf{93.07} & 97.20  & \textbf{97.21}                                                          \\ \hline
no                    & 0.06                                                                         & 34.48  & 23.50                                                             & 44.29 & 40.37                                                           & 44.45 & \textbf{55.05}$^\ddagger$ & 65.19 & 66.34                                                       & \textbf{68.44} & 65.73 & 75.72  & \textbf{76.11}                                                          \\
sv                    & 0.07                                                                         & 27.74  & 21.83                                                             & 43.83 & 29.66                                                           & 39.66 & \textbf{56.92}$^\ddagger$ & 74.84 & 76.38                                                       & \textbf{76.47} & 75.75 & 85.02  & \textbf{85.48}                                                          \\
fr                    & 0.09                                                                         & 28.92  & 22.29                                                             & 50.70 & 40.16                                                           & 47.13 & \textbf{62.72}$^\ddagger$ & 76.06 & 77.27                                                       & \textbf{79.62} & 79.24$^\ddagger$ & 88.57  & \textbf{88.82}                                                          \\
pt                    & 0.09                                                                         & 41.34  & 29.71                                                             & 59.00 & 48.78                                                           & 55.14 & \textbf{66.77} & 83.23 & 83.94                                                       & \textbf{84.94} & 84.64 & 90.66  & \textbf{91.10}                                                          \\
da                    & 0.10                                                                         & 51.77  & 37.61                                                             & 49.14 & 49.18                                                           & 56.87 & \textbf{63.21}$^\ddagger$ & 77.69 & 78.77                                                       & \textbf{79.53} & 79.02 & 87.19  & \textbf{87.61}                                                          \\
es                    & 0.12                                                                         & 41.86  & 31.49                                                             & 51.82 & 44.58                                                           & 50.82 & \textbf{60.29} & 76.58 & \textbf{79.43}$^\ddagger$                                                       & 77.93 & 77.91 & 86.56  & \textbf{86.88}                                                          \\
it                    & 0.12                                                                         & 37.14  & 24.65                                                             & 54.78 & 42.88                                                           & 49.93 & \textbf{66.55}$^\ddagger$ & 73.59 & \textbf{77.46}$^\ddagger$                                                       & 77.21 & 77.38$^\ddagger$ & 88.98  & \textbf{89.85}$^\ddagger$                                                          \\
hr                    & 0.13                                                                         & 29.35  & 22.11                                                             & 45.88 & 38.85                                                           & 45.87 & \textbf{55.40}$^\ddagger$ & 68.56 & 71.12$^\ddagger$                                                       & \textbf{71.70} & 71.08$^\ddagger$ & 82.57  & \textbf{83.45}$^\ddagger$                                                          \\
ca                    & 0.13                                                                         & 38.20  & 26.02                                                             & 50.97 & 46.24                                                           & 49.66 & \textbf{59.56} & 74.61 & \textbf{78.22}$^\ddagger$                                                       & 75.84 & 77.96$^\ddagger$ & 85.85  & \textbf{86.11}                                                          \\
pl                    & 0.13                                                                         & 43.13  & 31.41                                                             & 49.16 & 32.72                                                           & 52.02 & \textbf{62.97}$^\ddagger$ & 66.93 & \textbf{68.73}                                                       & 65.82 & 68.53 & 80.11  & \textbf{80.61}                                                          \\
uk                    & 0.13                                                                         & 29.60  & 28.99                                                             & 42.70 & 35.14                                                           & 49.40 & \textbf{56.13}$^\ddagger$ & 73.45 & \textbf{75.15}                                                       & 73.23 & 75.08 & 83.83  & \textbf{84.41}                                                          \\
sl                    & 0.13                                                                         & 33.67  & 29.08                                                             & 43.94 & 40.00                                                           & 42.80 & \textbf{60.01}$^\ddagger$ & 62.76 & 64.68                                                       & \textbf{67.42} & 64.49 & 75.71  & \textbf{76.58}                                                          \\
nl                    & 0.14                                                                         & 34.95  & 24.08                                                             & 48.69 & 34.87                                                           & 45.02 & \textbf{63.30}$^\ddagger$ & 79.28 & 79.07                                                       & 79.48 & \textbf{80.08} & 86.90  & \textbf{87.68}                                                          \\
bg                    & 0.14                                                                         & 28.42  & 24.71                                                             & 39.78 & 29.98                                                           & 43.98 & \textbf{54.49}$^\ddagger$ & 68.63 & \textbf{70.13}                                                       & 68.93 & 69.42 & \textbf{80.66}  & 80.60                                                          \\
ru                    & 0.14                                                                         & 29.65  & 26.83                                                             & 47.96 & 40.12                                                           & 51.53 & \textbf{59.20}$^\ddagger$ & 78.39 & 79.53                                                       & 77.88 & \textbf{79.76} & 88.80  & \textbf{89.05}                                                          \\
de                    & 0.14                                                                         & 31.52  & 25.12                                                             & 43.24 & 35.01                                                           & 41.11 & \textbf{50.59} & 68.41 & 67.75                                                       & \textbf{69.00} & 68.96 & \textbf{81.02}  & 80.66                                                          \\
he                    & 0.14                                                                         & 21.67  & 19.74                                                             & 34.82 & 24.38                                                           & 33.97 & \textbf{39.24} & 64.30 & 66.12                                                       & 65.22 & \textbf{66.88}$^\ddagger$ & 69.74  & \textbf{70.30}                                                          \\
cs                    & 0.14                                                                         & 34.06  & 27.53                                                             & 48.30 & 36.97                                                           & 51.00 & \textbf{63.79}$^\ddagger$ & 75.05 & \textbf{76.46}                                                       & 75.17 & 76.17 & 85.71  & \textbf{86.20}                                                          \\
ro                    & 0.15                                                                         & 35.05  & 28.36                                                             & 45.43 & 41.10                                                           & 46.29 & \textbf{61.23}$^\ddagger$ & 69.17 & \textbf{72.42}$^\ddagger$                                                       & 71.93 & 71.37$^\ddagger$ & 81.38  & \textbf{82.03}                                                          \\
sk                    & 0.17                                                                         & 39.98  & 35.62                                                             & 47.66 & 38.46                                                           & 50.28 & \textbf{61.10}$^\ddagger$ & 68.13 & \textbf{69.21}                                                       & 68.77 & 68.46 & 80.50  & \textbf{81.23}                                                          \\
id                    & 0.17                                                                         & 32.54  & 24.09                                                             & 34.11 & 26.76                                                           & \textbf{41.61} & 39.10 & 59.97 & \textbf{62.08}$^\ddagger$                                                       & 60.13 & 62.07$^\ddagger$ & 71.80  & \textbf{72.86}$^\ddagger$                                                          \\
lv                    & 0.18                                                                         & 49.05  & 35.21                                                             & 52.43 & 50.66                                                           & \textbf{58.80} & 57.87 & 64.74 & \textbf{68.46}$^\ddagger$                                                       & 66.81 & 67.94$^\ddagger$ & 79.01  & \textbf{79.64}                                                          \\
fi                    & 0.20                                                                         & 37.88  & 29.52                                                             & 44.83 & 38.53                                                           & 47.16 & \textbf{53.85} & 68.65 & \textbf{72.04}$^\ddagger$                                                       & 69.22 & 71.87$^\ddagger$ & 81.09  & \textbf{81.98}$^\ddagger$                                                          \\
et                    & 0.20                                                                         & 30.13  & 26.50                                                             & 41.86 & 25.64                                                           & 42.00 & \textbf{49.81} & 57.50 & \textbf{63.00}$^\ddagger$                                                       & 58.27 & 62.29$^\ddagger$ & 75.66  & \textbf{75.80}                                                          \\
zh                    & 0.23                                                                         & 26.66  & 23.95                                                             & 28.92 & 24.81                                                           & 27.94 & \textbf{31.20} & 53.37 & 55.58$^\ddagger$                                                       & 53.51 & \textbf{55.66}$^\ddagger$ & 63.86  & \textbf{64.06}                                                          \\
ar                    & 0.26                                                                         & 7.75   & 12.40                                                             & \textbf{25.77} & 4.32                                                            & 24.97 & 22.29 & 25.94 & \textbf{29.96}$^\ddagger$                                                       & 24.39 & 29.01$^\ddagger$ & 27.68  & \textbf{36.94}$^\ddagger$                                                          \\
la                    & 0.28                                                                         & -      & -                                                                 & -     & -                                                               & -     & -     & 42.22 & \textbf{46.11}$^\ddagger$                                                       & 43.32 & 43.34 & 45.52  & \textbf{46.96}$^\ddagger$                                                          \\
ko                    & 0.33                                                                         & 10.79  & 7.87                                                              & \textbf{21.82} & 5.16                                                            & 16.71 & 19.02 & 35.77 & 38.17$^\ddagger$                                                       & 36.08 & \textbf{39.97}$^\ddagger$ & 42.18  & \textbf{45.45}$^\ddagger$                                                          \\
hi                    & 0.40                                                                         & 20.97  & 17.68                                                             & 27.66 & 25.11                                                           & 29.00 & \textbf{35.05} & 47.95 & \textbf{52.76}$^\ddagger$                                                       & 49.45 & 51.90$^\ddagger$ & \textbf{56.45}  & 55.63                                                          \\
ja                    & 0.49                                                                         & -      & -                                                                 & -     & -                                                               & -     & -     & 40.59 & \textbf{43.17}$^\ddagger$                                                       & 41.46 & 42.82$^\ddagger$ & 44.97  & \textbf{45.08}                                                          \\ \hline
avg                   & 0.17                                                                         & 32.44  & 25.64                                                             & 43.55 & 34.66                                                           & 44.11 & \textbf{53.10} & 64.72 & \textbf{66.98}                                                       & 65.91 & 66.83 & 75.12  & \textbf{75.97}                                                          \\ \hline
\end{tabular}
}
\caption{Zero-shot cross-lingual results on the POS task (averaged over three runs). Languages are sorted by the word-ordering distance to English. Since the word-level embeddings for la and ja languages are absent, we do not report these results. We use `$\ddagger$' to denote the performance improvement of the proposed models is higher than their corresponding average improvements.}
\label{pos_results}
\end{table*}

\section{Experiments}

\subsection{Datasets}
We test our methods on three sequence labeling tasks in the cross-lingual setting, namely, part-of-speech tagging (POS), named entity recognition (NER), and slot filling (SF). 
For the POS task, we choose the same language set as~\citet{ahmad2018difficulties} (31 languages in total) from the Universal Dependencies~\cite{nivre2017universal} to evaluate our methods.
% choose English (en), French (fr), Spanish (es), Portuguese (pt), Greek (el), and Russian (ru) to evaluate our methods. 
And we use the CoNLL 2002 and CoNLL 2003 datasets~\cite{tjong2002introduction,sang2003introduction}, which contain English (en), German (de), Spanish (es), and Dutch (nl), to evaluate our methods for the NER task.
Finally, for the SF task, we use the multilingual natural language understanding (NLU) (containing the intent detection and slot filling tasks) dataset introduced by~\citet{schuster2019crosslingual}, which contains English (en), Spanish (es), and Thai (th) across weather, alarm and reminder domains. The data statistics for these datasets are in the appendix.

\subsection{Experimental Settings}
\subsubsection{Our Models and Baselines}
All our models and baseline models consist of a sequence encoder to produce features for input sequences and a conditional random field (CRF) layer~\cite{lample2016neural,ma2016end} to make predictions based on the sequence features. 
For the sequence encoder, we use Bidirectional LSTM (\textbf{BiLSTM}), Transformer (\textbf{TRS}) using sinusoidal functions as positional embeddings, Relative Positional Transformer (\textbf{RPT}) proposed in~\citet{ahmad2018difficulties}\footnote{They utilize the relative positional embeddings proposed in~\citet{shaw2018self} to encode less word order information for cross-lingual adaptation.}, or Order-Reduced Transformer (\textbf{ORT}).
All Transformer-based encoders use Conv1d as the feed-forward layer for a fair comparison.
Word order shuffling is applied to BiLSTM and TRS baselines to make them less sensitive to word orders.
We fine-tune M-BERT by adding a linear layer on top of it, and we compare two different fine-tuning M-BERT methods (with and without freezing the positional embeddings).

\subsubsection{Training Details}
We evaluate our models with cross-lingual word embeddings (word-level) and M-BERT embeddings (subword-level). For the word-level embeddings, we leverage RCSLS~\cite{joulin2018loss} for the POS and NER tasks, and we use the refined RCSLS in~\citet{liu2019zero} for the SF task since it is specifically refined for this task. 
% and M-BERT~\cite{devlin2019bert} for the zero-shot adaptation.
% We freeze the cross-lingual embeddings of RCSLS and M-BERT (i.e., freeze the parameters of M-BERT) in the training phase, and our model can directly adapt to target languages in the test phase.
% We freeze RCSLS embeddings in the training phase to ensure the cross-lingual alignment between the source and target languages.
% We try two settings for the M-BERT. First, same as what we do to RCSLS, we freeze the M-BERT embeddings and add sequence labeling models on top of M-BERT.
We set the kernel size as 3 for the feed-forward layer Conv1d in the Transformer encoder. For the word order shuffled data, we generate ten different word order shuffled samples with $k=\infty$ (can generate any permutation) for each source language training sample.
Note that the word order shuffling can not be applied for M-BERT-based models since they are pre-trained based on the correct language order, and it is not suitable to feed them with order-shuffled sequences.
For all the tasks, we use English as the source language and other languages as target languages. We follow~\citet{ahmad2018difficulties} to calculate the language distances between target languages and English.
% In the zero-shot scenario, we do not use any data samples in the target languages, while in the few-shot setting, we utilize a few training samples in the target languages. 
% In this paper, we mainly focus on the effectiveness of our methods on the zero-shot adaptation, and we also explore the effectiveness of our methods in the 
% performance changes over different numbers of target language training samples.
We use the standard BIO-based F1-score for evaluating the NER and SF tasks, as in~\citet{lample2016neural}, and accuracy score for evaluating the POS task, as in~\citet{kim2017cross}. 
% And for the NLU task, we only take the slot filling task for the investigation of sequence labeling and remove the intent detection task. 
More details are in the appendix.

\subsubsection{Applying ORT into Strong Baselines}
We apply the ORT into two strong cross-lingual models for zero-shot cross-lingual NLU~\cite{liu2019zero} and NER~\cite{chen2019multi}, and both are based on BiLSTM as the sequence encoder, which is order-sensitive. To ensure a fair comparison, we keep all settings as in the original papers, except that the sequence encoder is replaced. 
For the NLU model from~\citet{liu2019zero}, we replace the BiLSTM with ORT. And for the NER model from~\citet{chen2019multi}, we replace the BiLSTM in the shared feature extractor module with ORT.

\section{Results \& Discussion}
\subsection{Zero-shot Adaptation}
% In this section, we follow several questions to analyze the zero-shot adaptation results.
% Question \#1: Do order-reduced models improve cross-lingual performance?
% Question \#2: Does removing the CRF layer improve cross-lingual performance?
% Question \#3: How does Performance improvements relate to language distances. ORT: Close languages have close local order information. M-BERT PE tends to give better performance to distant languages.

% \subsubsection{Can order-reduced Models Improve Cross-lingual Performance?}
% \subsubsection{order-reduced Models}

\subsubsection{Order-Reduced Transformer}
% 1. Describe improvements in target languages made by our methods
% 2. Apply our approach to the SOTA model
% In general, order-agnostic models (i.e., Transformer based models without positional embeddings) outperform their corresponding vanilla Transformer, Transformer with relative positional embeddings~\cite{ahmad2018difficulties} and commonly used BiLSTM+CRF structure~\cite{lample2016neural,schuster2019crosslingual}.
As we can see from Table~\ref{zero_shot_results}, removing positional embeddings from Transformer (ORT) only makes the performance in the source language (English) drop slightly (around 0.5\%). This indicates that leveraging only local order information results in a good performance in sequence labeling tasks. In other words, relying just on the information from the neighboring words (how many neighboring words depend on the kernel size in Conv1d) can ensure relatively good performance for sequence labeling tasks.
On the other hand, in terms of zero-shot adaptation to target languages (from Table~\ref{zero_shot_results} and~\ref{pos_results}), ORT achieves consistently better performance than the order-sensitive encoders (i.e., BiLSTM and TRS) as well as the order-reduced encoder RPT~\cite{ahmad2018difficulties}. 
For example, in the SF task, in terms of the average performance of using word-level embeddings, ORT outperforms BiLSTM, TRS, and RPT by 6.12\%, 3.69\%, and 3.58\% on the F1-score, respectively. 

Compared to the order-sensitive models, ORT fits the word order of the source language less, which increases its adaptation robustness to target languages. We conjecture that the reason why ORT outperforms RPT is that RPT still keeps the relative word distances. Although RPT reduces the order information that the model encodes, it might not be suitable for target languages that do not have similar relative word distance patterns to English, while ORT removes all the order information in positional embeddings, which makes it more robust to the word order differences.

\subsubsection{Shuffling Word Order}
From Table~\ref{zero_shot_results} and~\ref{pos_results}, we can see that the models trained with word order shuffled data lead to a visible performance drop in English, especially for the POS and NER tasks. For target languages, however, we observe that the performance improves in the SF task by using such data. For example, using the order shuffled data to train the Transformer improves the performance by 1.52\% on the averaged F1-score.
For cross-lingual adaptation, performance loss in the source language has a negative impact on the performance in target languages. In the SF task, the performance drop in English is relatively small ($\sim$2\%); hence, the benefits from being less sensitive to word orders are greater than the performance losses in English.

On the other hand, for the NER and POS tasks, using order shuffled data makes the performance in target languages worse. For example, for the POS task, the average accuracy drops 8.89\% for the Transformer trained with the order shuffled data compared to the one trained without such data.
We observe large performance drops for the NER and POS tasks in English caused by using the order shuffled data (for example, for the POS task, the drop is $\sim$6\%) since the models for these tasks are more vulnerable to the shuffled word order. In this case, the performance losses in English are larger than the benefits of being less sensitive to word orders.

\subsubsection{Order-Agnostic Positional Embeddings}
% like investigation
As we can see from Table~\ref{zero_shot_results} and~\ref{pos_results}, compared to TRS, we observe that TRS trained with M-BERT PE (frozen) only results in a slight performance drop in English, while it generally brings better zero-shot adaptation performance to target languages. For example, in the POS task, TRS with M-BERT PE achieves 2.26\% higher averaged accuracy than the one without M-BERT PE.
Since M-BERT is trained using 104 languages, positional embeddings in M-BERT are fitted to different word orders across various languages and become order-agnostic. 
% The frozen M-BERT positional embeddings still provide order information in the training phase so that the model maintains similar performance in English. And 
Since the pre-trained positional embeddings are frozen, their order-agnostic property is retained, which brings more robust adaptation to target languages.

In addition, we notice that ORT achieves similar performance to TRS with M-BERT PE, which further illustrates the effectiveness of encoding partial order information for zero-shot cross-lingual adaptation.

\begin{table}[t]
\centering
\resizebox{0.42\textwidth}{!}{
\begin{tabular}{lcccc}
\hline
\multicolumn{1}{l|}{\multirow{2}{*}{}}  & \multicolumn{2}{c|}{\textbf{Spanish}}                           & \multicolumn{2}{c}{\textbf{Thai}}              \\ \cline{2-5} 
\multicolumn{1}{l|}{}     & \textbf{ID}                    & \multicolumn{1}{c|}{\textbf{SF}}  & \textbf{ID}                    & \textbf{SF}                      \\ \hline
\multicolumn{1}{l|}{\citet{liu2019zero}}    & 90.20                     & \multicolumn{1}{c|}{65.79} & 73.43                     & 32.24  \\ 
\multicolumn{1}{l|}{\quad using TRS}    & 89.71                     & \multicolumn{1}{c|}{67.10} & 74.68                     & 31.20  \\
\multicolumn{1}{l|}{\quad using ORT}    & \textbf{91.46}                     & \multicolumn{1}{c|}{\textbf{71.36}} & \textbf{75.02}                     & \textbf{34.61}  \\ \hline
\end{tabular}
}
\caption{Zero-shot results for the intent detection (ID) accuracy and SF F1-score on the NLU task.}
\label{NLU_SOTA}
\end{table}

\begin{table}[t]
\centering
\resizebox{0.42\textwidth}{!}{
\begin{tabular}{lcccc}
\hline
\multicolumn{1}{l|}{}  & \textbf{de}    & \textbf{es}                        & \multicolumn{1}{c|}{\textbf{nl}}    & \textbf{avg}                   \\ \hline
\multicolumn{1}{l|}{\citet{chen2019multi}}  & 56.00                     & 73.50                     & \multicolumn{1}{c|}{72.40} & 67.30                     \\ 
\multicolumn{1}{l|}{\quad using TRS}  & 56.89                     & 73.72                     & \multicolumn{1}{c|}{72.22} & 67.61                     \\ 
\multicolumn{1}{l|}{\quad using ORT}  & \textbf{58.97}                     & \textbf{74.65}                     & \multicolumn{1}{c|}{\textbf{72.56}} & \textbf{68.73}           \\ \hline
\end{tabular}
}
\caption{Zero-shot results on the NER task.}
\label{NER_SOTA}
\end{table}

\subsubsection{Fine-tuning M-BERT}
From Table~\ref{zero_shot_results} and~\ref{pos_results}, we observe that the results of fine-tuning M-BERT in the source language, English, are similar for both methods (less than 0.1\% difference) while freezing the positional embeddings in the fine-tuning stage generally brings better zero-shot cross-lingual performance in target languages.
% when fine-tuning M-BERT generally achieves better zero-shot performance in target languages than the original fine-tuning approach. 
Although positional embeddings are frozen, they can still provide order information for the model to encode sequences, which ensures the performance in English does not greatly drop. In the meantime, the positional embeddings are not affected by the English word order, and the order-agnostic trait of the positional embeddings is preserved, which boosts the generalization ability to target languages.

\begin{table}[t!]
\centering
\resizebox{0.3\textwidth}{!}{
\begin{tabular}{lccc}
\hline
\multicolumn{1}{l|}{}  & $k=0$ & $k=1$   & $k=2$  \\ \hline
\multicolumn{1}{r|}{BiLSTM} & \textbf{94.87} & 85.16    & 83.68  \\ 
% \multicolumn{1}{r|}{BiLSTM w/ shuffled data}                     & \textbf{93.66}  \\
% \multicolumn{1}{l|}{Transformer+Linear}  & 92.27                     & 82.31  \\ 
% \multicolumn{1}{l|}{\quad w/ order shuffled data}  & 98.26                     & 91.47  \\
\multicolumn{1}{r|}{TRS}    &94.78 & 84.56               & 83.06  \\
% \multicolumn{1}{r|}{Transformer w/ shuffled data}                   & 92.70  \\
% \multicolumn{1}{l|}{ORT+Linear}  & 97.35                     & 86.34  \\ 
\multicolumn{1}{r|}{RPT} 
& 94.23 & 84.93 & 83.86  \\ \hline
\multicolumn{1}{r|}{ORT}   &94.50 & \textbf{87.87}          & \textbf{86.95}  \\ \hline
\end{tabular}
}
\caption{Slot F1-scores on different noisy SF test sets in English. $k=0$ denotes the original English test set.
% using refined RCSLS embeddings from~\citet{liu2019zero}.
}
\label{nlu_shuffle_En}
\end{table}

\subsubsection{Applying ORT to Strong Models}
As shown in Table~\ref{NLU_SOTA} and~\ref{NER_SOTA}, we leverage ORT to replace the order-sensitive encoder, BiLSTM, in the strong zero-shot cross-lingual sequence labeling models proposed in~\citet{liu2019zero} and \citet{chen2019multi}. The zero-shot cross-lingual NLU model proposed in~\citet{liu2019zero} is the current state-of-the-art for the multilingual NLU dataset~\cite{schuster2019crosslingual}, and the model in~\citet{chen2019multi} achieves promising results in the zero-shot cross-lingual NER task. As we can see, replacing the order-sensitive encoders in their models with ORT can still boost the performance. We conjecture that since there are always cross-lingual performance drops caused by word order differences, reducing the word order of the source language fitted into the model can improve the performance.

In addition, we observe that the performance stays similar when we replace BiLSTM with TRS, which illustrates that the performance improvement made by ORT does not come from TRS, but from the model's insensitivity to word order.

\subsubsection{Improvements vs. Language Distance}
As we can see from Table~\ref{pos_results}, our proposed order-reduced models (e.g., ORT and TRS w/ M-BERT PE) outperform baseline models in almost all languages.
% , including Spanish (es) and French (fr), which have a close language distance to English, and Greek (el) and Thai (th), which are lexically and syntactically different from English.
% 1. Word-level Embeddings
In the word-level embeddings setting, we observe that languages that are closer to English benefit more from ORT, since most numbers denoted with `$\ddagger$' come from languages that are close to English. We conjecture that ORT predicts the label of a token based on the local information of this token (the token itself and its neighbor tokens), and languages that are closer to English could have a more similar local word order to English. Interestingly, for M-BERT fine-tuning and models using M-BERT embeddings, we can see that languages that are further from English benefit more from order-reduced models.
We speculate that the alignment quality of topologically close languages in M-BERT is generally good, and the cross-lingual performance for the languages that are closer to English is also overall satisfactory, which narrow down the improvement space for these languages. In contrast, the performance boost for languages that are further from English becomes larger. Surprisingly, freezing PE in the M-BERT fine-tuning significantly improves the performance of Arabic (ar). Given that the cross-lingual performance of the original M-BERT fine-tuning in Arabic is relatively low, we conjecture that one of the major reasons comes from the word order discrepancy between English and Arabic, which leads to a large performance improvement made by freezing the M-BERT PE.

% 2. M-BERT Embeddings
% This is because, although different languages naturally have different word orders, they are more likely to have the same word orders in some local areas, such as the entity names. Hence, ORT, which only encodes local word order information, generally improves performance when transferring to close and distant target languages compared to the other baselines.
% As for Transformer with frozen M-BERT positional embeddings, it can also generally boost performance on different target languages given the generalization ability of the order-agnostic positional embeddings.
% and the baseline models contain or partially contain the source language order information. 
% Order-agnostic models are able to learn the task in the source language well enough due to the extensive training data and have a better adaptation ability to target languages since the models do not overfit to the source language word order.

% TRS vs BiLSTM (NLU BiLSTM is better)
% In addition, we observe that TRS+CRF generally achieves better results than BiLSTM+CRF. We conjecture that it is because BiLSTM contains more order information than the Transformer since BiLSTM has a memory cell to remember the previous tokens, and Transformer only leverages positional embeddings for modeling word orders. Hence, BiLSTM might have a more serious overfitting problem to the source language word order.

% \subsubsection{Does Our Model Really Insensitive to Order Differences?}
\subsubsection{How Order-Insensitive Is Our Model?}
% To test how our proposed model are order-reduced, we follow~\cite{lample2017unsupervised} to slightly shuffle the word order of the sequences to create a noisy English test set. Concretely, we first keep the order of tokens in each entity the same and consider them as one ``word''. Then we randomly select some words to replace with their neighbor words within two-word distances.
To test the word order insensitivity of ORT, we follow the order shuffled methods in Section~\ref{shuffle_sec}, and set $k=1$ and $k=2$ to slightly shuffle the word order of the sequences and create a noisy English test set. As we can see from Table~\ref{nlu_shuffle_En}, ORT achieves better results than BiLSTM, TRS, and RPT on the noisy SF test set, which further illustrates that ORT is more insensible and resistant to word order differences than the baseline encoders. This property improves the generalization ability of ORT to target language word orders.
% \footnote{We do not include the shuffled word order and M-BERT based models in the table. For the former one, it is not fair to compare baselines with the models trained with word order shuffled data since the training set has a similar distribution to the test set. For the latter one, the M-BERT-based models are pre-trained based on the correct language order. Hence, it is not suitable to feed them with the order-shuffled test set.}
% Additionally, models trained with order shuffled data are the most resistant to the noisy input sequences. This is because, in the training phase, the models leverage equivalent or more noisy training samples, which makes them more robust to order differences.

\begin{figure}[t!]
\begin{subfigure}{0.45\textwidth}
    \centering
    \includegraphics[scale=0.34]{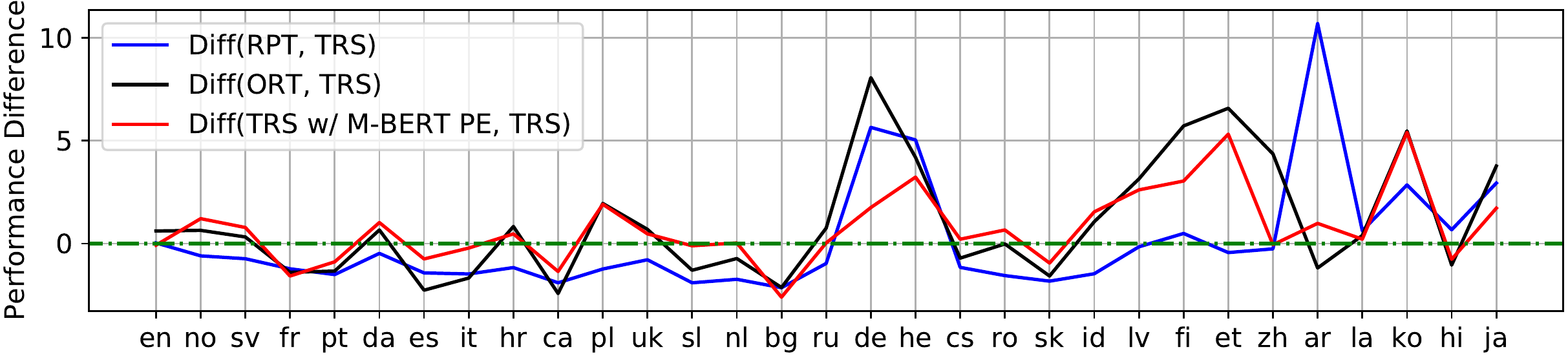}
    \caption{Performance difference on noun.}
\end{subfigure}
\begin{subfigure}{0.45\textwidth}
    \centering
    \includegraphics[scale=0.34]{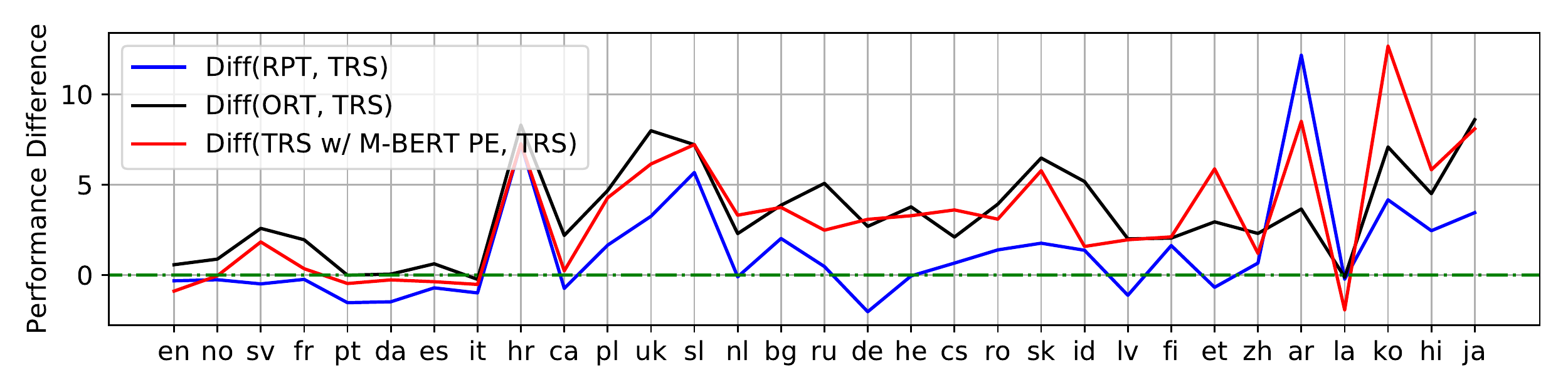}
    \caption{Performance difference on verb.}
\end{subfigure}
\caption{Analysis on specific part-of-speech types. Languages are sorted by the word-ordering distance to English. We use Diff(A, B) to denote how much A outperforms B.}
\label{fig:performance_by_type}
\end{figure}

\subsubsection{Performance Breakdown by Types}
We compare different models (based on the frozen M-BERT embeddings) on specific part-of-speech types for the POS task. From Figure~\ref{fig:performance_by_type}, we observe that, in general, languages that are further from English benefit more from our proposed order-reduced models, which accords with the findings from Table~\ref{pos_results}. Interestingly, we find that the improvements of order-reduced models (ORT and TRS w/ M-BERT PE) on the verb are larger than on the noun. 
We speculate that the word orders of the verb's surrounding words are different across languages, while the set of its surrounding words is more likely to remain the same or similar across languages at the semantic-level, which boost the advantages of our proposed models, especially for the ORT which relies greatly on the extracted n-gram features from the neighbor words for the prediction.
Additionally, we find that the improvement made by RPT is also more significant on the verb than on the noun. We conjecture that the verb's relative positions with other part-of-speech types are more similar across languages compared to that for the noun. The experiments on more part-of-speech types are shown in the appendix.

\subsection{Few-shot Adaptation}
\begin{figure}[t!]
\begin{subfigure}{0.45\textwidth}
    \centering
    \includegraphics[scale=0.58]{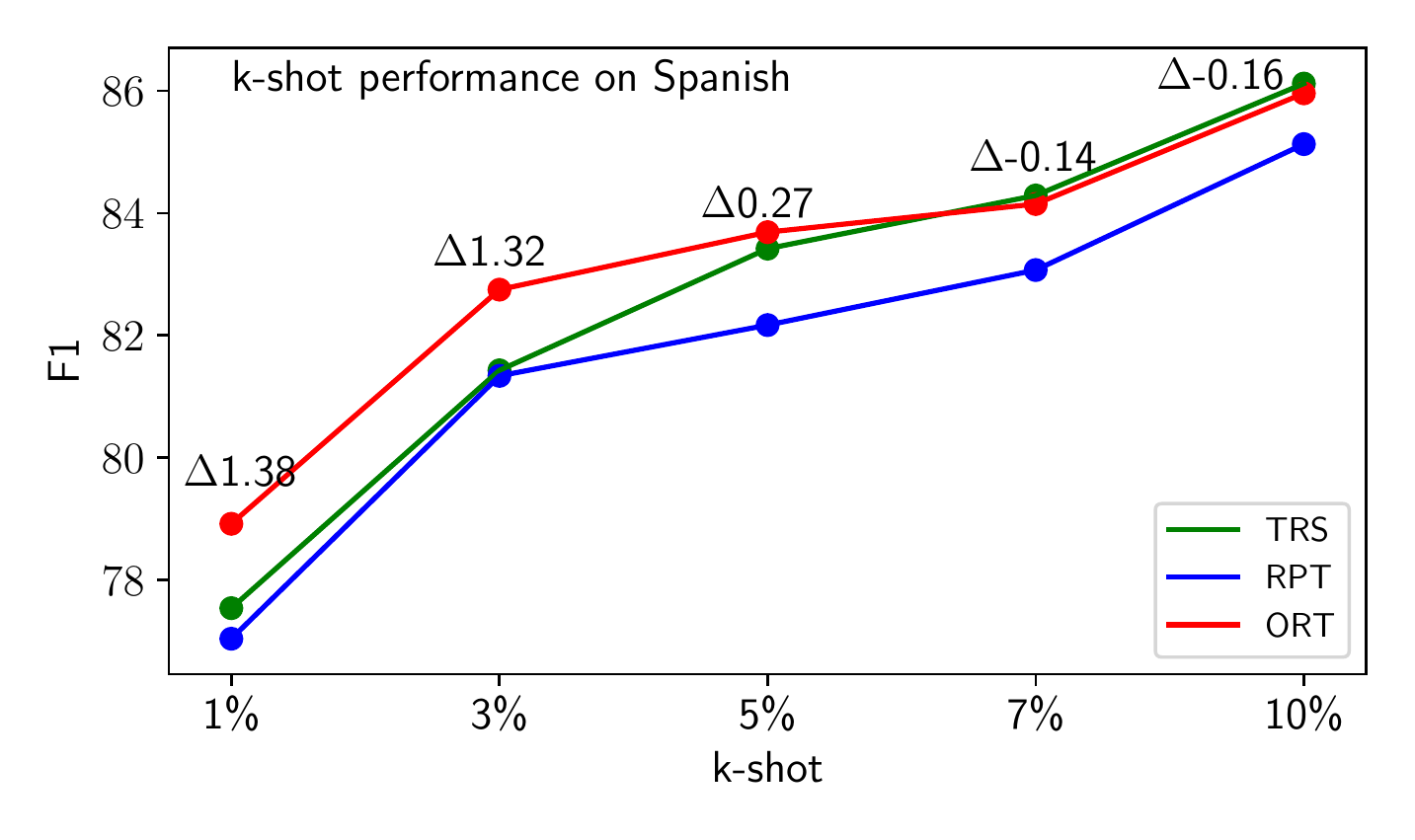}
\end{subfigure}
\begin{subfigure}{0.45\textwidth}
    \centering
    \includegraphics[scale=0.58]{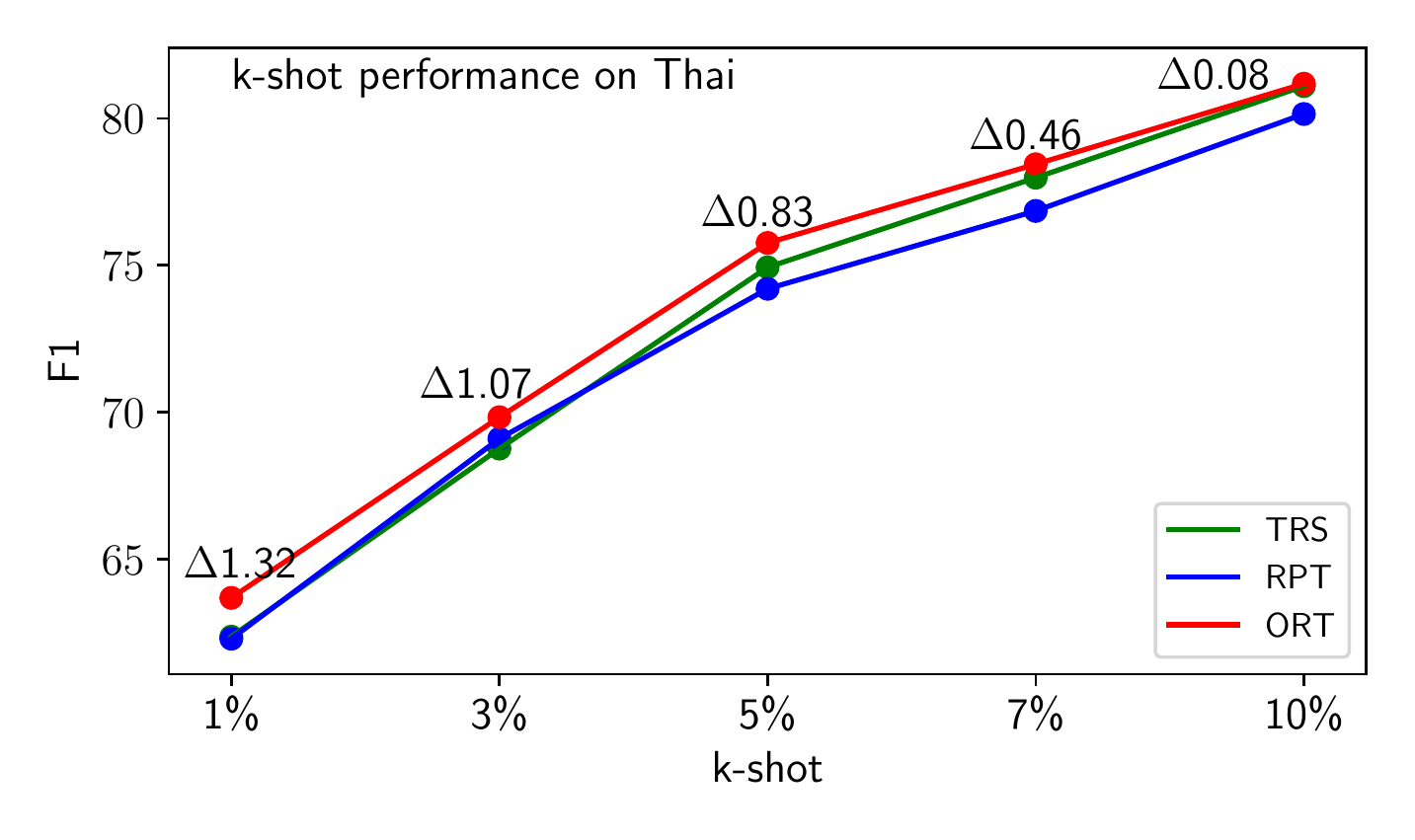}
\end{subfigure}
\caption{Few-shot F1-scores for the SF task for Spanish and Thai. The x-axis represents the proportion of target language training samples in the training set. The numbers with $\Delta$ denote how much ORT outperforms TRS.}
\label{fig:fewshot}
\end{figure}

Since we do not observe the order information for target languages in the zero-shot scenario, the order-reduced models will have a more robust adaptation ability. Then, the question we want to ask is whether order-reduced models can still improve the performance if a few training samples in target languages are available. 
We test with different numbers of target language training samples for the SF task, and the results are shown in Figure~\ref{fig:fewshot}. 
We observe that the improvements in the few-shot scenarios are lower than the zero-shot scenario, where ORT improves TRS by 4.17\% and 3.22\% on Spanish and Thai, respectively (from Table~\ref{zero_shot_results}), and as the proportion of target language training samples goes up, the improvement made by ORT goes down. This is because the model is able to learn the target language word order based on the target language training samples, which decreases the advantages of the order-reduced models. 
We also observe that RPT generally achieves worse performance than TRS, and we conjecture that RPT requires more training samples to learn the relative word order information than TRS, which lowers its generalization ability to the target language in the few-shot scenario.

\subsection{Ablation Study}
In this section, we explore the model variations in terms of positional embeddings, the feed-forward layer for TRS and ORT, adding different permutations to the shuffled word order, and whether to use the CRF layer.
We test the models' zero-shot performance on the SF task for this ablation study, and results are illustrated in Table~\ref{ablation_study}. 

% positional embeddings (trainable and sinusoidal)
\subsubsection{Positional Embeddings}
We observe that TRS+CRF using trainable positional embeddings achieves similar performance to using sinusoidal positional embeddings. 

% Feed-forward Layer
\subsubsection{Feed-forward Layer}
We can see that the performance of TRS+CRF using linear layers as the feed-forward layer is on par with using Conv1d. We conjecture that the reason is that the positional embeddings in TRS have already encoded the word order information of the whole input sequence which makes the type of feed-forward layer less important.
% We conjecture that the reason is that using Conv1d encodes extra source language word order information compared to using the linear layer, which decreases the generalization ability to target languages.
However, when we replace Conv1d with linear layers for the feed-forward layer in ORT+CRF, the performance greatly drops ($\sim$8.5\% F1-score drops for Spanish and $\sim$5\% F1-score drops for Thai), and the performance continues to significantly drop when the CRF layer is removed (ORT+Linear).
This is because ORT can not encode any order information when the feed-forward layer  Conv1d is replaced with the linear layer, and not any word order information is injected into the ORT+Linear model in which the CRF layer is removed. This makes the model perform badly in the source language and then weakens its adaptation ability to target languages.
In addition, we observe that the Conv1d feed-forward layer is also important for TRS trained with order-shuffled data. This is because Conv1d encodes the order of tokens in the entity (we do not shuffle the tokens in an entity), which is essential for detecting entities.

\subsubsection{Kernel Size vs. Performance}
\begin{figure}[t!]
\centering
\includegraphics[scale=0.60]{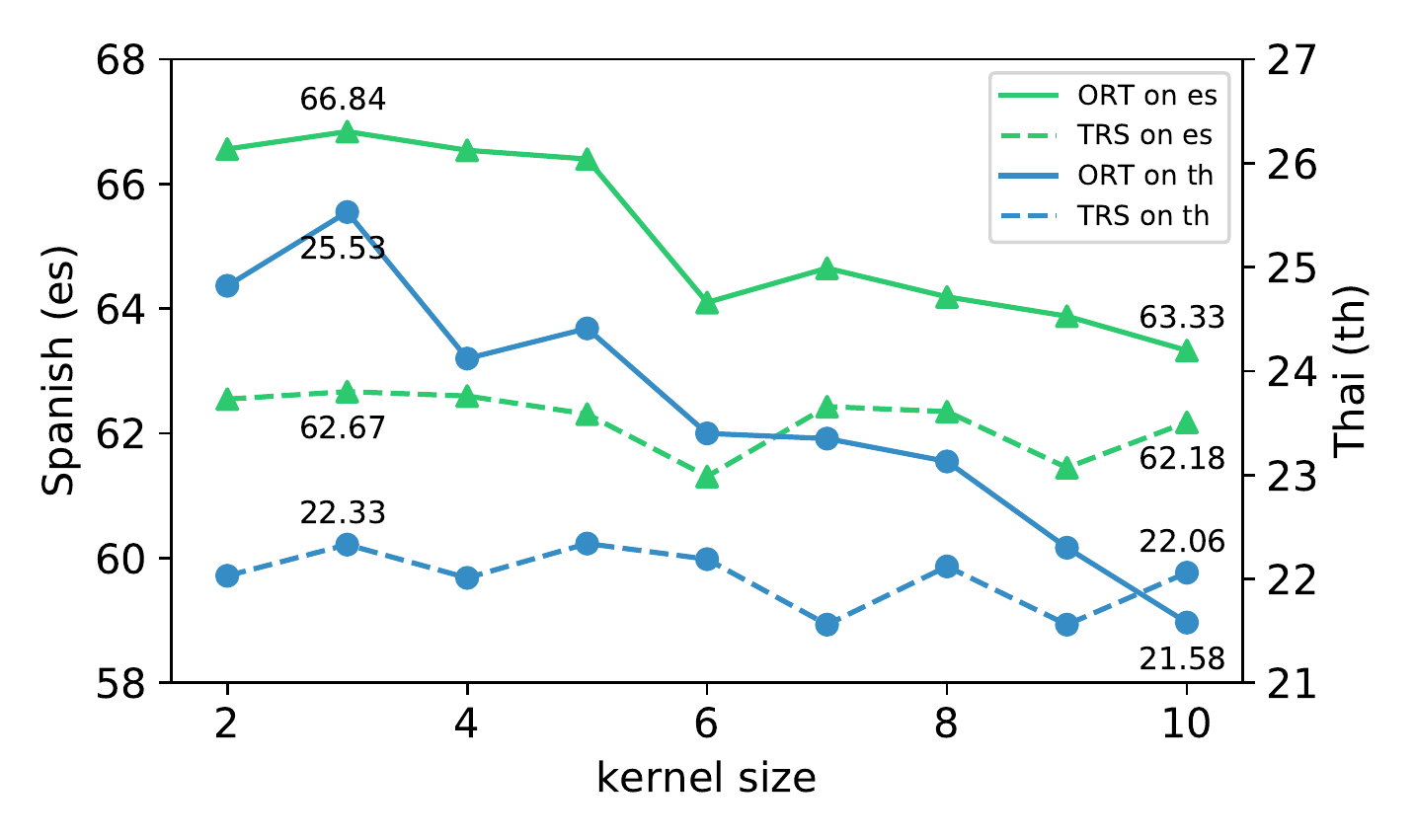}
\caption{Zero-shot results on the SF task with different kernel sizes for ORT and TRS.}
\label{fig:kerne_size}
\end{figure}

% we test ORT with different kernel sizes of Conv1d.
%, which represent different amounts of local order information.
Since the kernel size of Conv1d represents the amounts of local word order information that ORT encodes, we explore how the kernel size affects the performance.
As shown in Figure~\ref{fig:kerne_size}, with the increase of kernel size, the zero-shot performance of ORT decreases, and the performance of ORT becomes similar to TRS's when the kernel size is 10. This is because the larger the kernel size, the more order information the model will encode. Hence, the model's generalization ability to target languages decreases when the kernel size is too large.

% order shuffled data (k)
\subsubsection{Different Permutations of Word Order}
We try using different permutations (changing the value of k) of the word order to generate order-shuffled data.
As we can see, when we slightly shuffle the word order ($k=2$), the performance becomes worse than not using order-shuffled data. This is because the model fits the slightly shuffled word order, which is not similar to the target languages. After more perturbations are added to word order, TRS becomes more robust to order differences.

\begin{table}[t!]
\centering
\resizebox{0.48\textwidth}{!}{
\begin{tabular}{lccc|cc}
\hline
\multicolumn{1}{l}{} & \textbf{PE}   & \textbf{Feed-forward}  & \textbf{k} &  \textbf{es}    & \textbf{th}    \\ \hline
% TRS+CRF           & Trainable & linear$^\ddagger$    & -  & 62.13 & 22.68 \\ 
% \quad w/ shuffled data          & Trainable & linear$^\ddagger$  & $\infty$    & 58.61 & 20.02 \\ 
% TRS+CRF           & Trainable & 3  & -    & 60.38 & 22.18 \\
% \quad w/ shuffled data  & Trainable & 3  & $\infty$    & 63.56 & 24.25 \\ \hline
TRS + CRF           & Trainable & Linear    & -  & 62.13 & 22.68 \\ 
TRS + CRF           & Sinusoid  & Linear & -     & 62.55 & 21.82 \\
\quad w/ shuffled data          & Sinusoid & Linear  &  $\infty$   & 58.89 & 19.27 \\ \hline
TRS + Linear           & Sinusoid  & Conv1d   & -    & 55.40 & 19.33 \\
TRS + CRF           & Sinusoid  & Conv1d   & -    & 62.67 & 22.33 \\
\quad w/ shuffled data  & Sinusoid & Conv1d   & 2   & 61.12 & 21.24 \\ 
\quad w/ shuffled data & Sinusoid & Conv1d   & 3   & 63.20 & 23.34 \\ 
\quad w/ shuffled data & Sinusoid & Conv1d   & 4   & 63.54 & 23.59 \\
\quad w/ shuffled data & Sinusoid & Conv1d   & $\infty$   & \textbf{63.86} & \textbf{24.17} \\ \hline
ORT + Linear         & -         & Linear & -     & 39.65 & 13.52 \\
ORT + CRF         & -         & Linear & -     & 58.27 & 20.35 \\ 
ORT + Linear         & -         & Conv1d & -     & 61.76 & 22.44 \\
% ORT + CRF         & -         & 2   & -   & 66.56 & 24.82 \\
ORT + CRF     & -         & Conv1d  & -    & \textbf{66.84} & \textbf{25.53} \\ \hline
% ORT + CRF         & -         & 5   & -   & 66.40 & 24.41 \\ 
% ORT + CRF         & -         & 7   & -   & 64.65 & 23.35 \\ 
% ORT + CRF         & -         & 8   & -   & 64.19 & 23.13 \\ 
% ORT + CRF         & -         & 10   & -   & 63.33 & 21.58 \\ 
\end{tabular}
}
\caption{Ablation study on positional embeddings, feed-forward layer, word order shuffling, and CRF layer. Results are the F1-scores for the zero-shot SF task.
% $^\ddagger$denotes the kernel size of the Conv1d. $^\ddagger$denotes that the model uses a linear layer as the feed-forward layer. k represents the tunable parameter for the order shuffle. 
``-'' denotes that the model does not have this module. ``+CRF'' and ``+Linear'' denotes using and not using the CRF layer, respectively.}
\label{ablation_study}
\end{table}

\subsubsection{Effectiveness of the CRF Layer}
For sequence labeling tasks, the CRF layer, which models the conditional probability of label sequences, could also implicitly model the source language word order in training. Therefore, we conduct an ablation study to test the effectiveness of the CRF layer for the cross-lingual models.
From Table~\ref{ablation_study}, we can see that removing the CRF layer makes the performance worse. We conjecture that although the CRF layer might contain some information on the word order pattern in the source language, 
%, which could hurt the performance of the model in target languages. 
It also models the conditional probability for tokens that belong to the same entity so that it learns when the start or the end of an entity is. This is important for sequence labeling tasks, and models that have the CRF layer removed might not have this ability.
For example, in the SF task, when the user says ``set an alarm for 9 pm'', ``for 9 pm'' belongs to the ``DateTime'' entity, and the CRF layer learns to model ``for'' and ``pm'' as the start and end of the ``DateTime'' entity, respectively. Without the CRF layer, models treat the features of these tokens independently.

\section{Conclusion}
% Word order differences naturally exist among different languages. 
% In this paper, we hypothesize that cross-lingual models that fit into the source language word order could fail to generalize to target language word orders. 
In this paper, we investigate whether reducing the word order of the source language fitted into the models can improve cross-lingual sequence labeling performance. We propose several methods to build order-reduced models, and then compare them with order-sensitive baselines.
Extensive experimental results show that order-reduced Transformer (ORT) is robust to the word order shuffled sequences, and it consistently outperforms the order-sensitive models as well as relative positional Transformer (RPT). Taking this further, ORT can also be applied to strong cross-lingual models and improve their performance. Additionally, preserving the order-agnostic property for the M-BERT positional embeddings gives the model a better generalization ability to target languages. 
% We also found out that there is a trade-off between maintaining the source language performance and insensitivity to word order. 
% We observe that the English performance of order-reduced models are usually lower than order-sensitive models
Furthermore, we show that encoding excessive or insufficient word order information leads to inferior cross-lingual performance, and models that do not encode any word order information perform badly in both source and target languages. 
% and cross-lingual sequence labeling models still need to preserve at least partial sensitivity to word order. 
% Models that encode too much word order information would have inferior cross-lingual performance, and models that do not encode any word order information perform badly in both source and target languages. 

% our proposed models are more robust to word order differences and have better zero-shot cross-lingual ability to target languages in sequence labeling tasks.
% In addition, we utilize this hypothesis to boost the performance upon a competitive model. First, we apply our proposed model to competitive cross-lingual models and achieve better performance. Second, we propose a better fine-tuning M-BERT approach.
% Furthermore, we conduct few-shot experiments and ablation studies to analyze order-reduced models further.

\section*{Acknowledgments}
We want to thank the anonymous reviewers for their helpful comments and constructive feedback. This work is partially funded by ITF/319/16FP and MRP/055/18 of the Innovation Technology Commission, the Hong Kong SAR Government.

\bibliography{aaai2021}

\end{document}